\documentclass[runningheads]{llncs}
\usepackage{soul,color,xcolor} 
\usepackage{graphicx}
\usepackage{multirow}
\begin{document}
\title{RobNODDI: Robust NODDI Parameter Estimation with Adaptive Sampling under Continuous Representation}
\titlerunning{Robust NODDI Parameter Estimation}
\author{Taohui Xiao\inst{1,4} \and
Jian Cheng\inst{2,3}\and
Wenxin Fan\inst{4}\and
Jing Yang\inst{4}\and
Cheng Li\inst{4}\and \\
Enqing Dong\inst{1}\and
Shanshan Wang\inst{4,5}}
\authorrunning{T. Xiao et al.}
%
\institute{School of Mechanical, Electrical \& Information Engineering, Shandong University, Weihai 264209, China \\
\email{enqdong@sdu.edu.cn} \and
State Key Laboratory of Software Development Environment, Beihang University, Beijing, China \and
Key Laboratory of Data Science and Intelligent Computing, Institute of International Innovation, Beihang University, Hangzhou, Zhejiang, China \and
Paul C. Lauterbur Research Center for Biomedical Imaging, Shenzhen Institute of Advanced Technology, Chinese Academy of Sciences, Shenzhen, Guangdong, China \\
\email{ss.wang@siat.ac.cn} \and
Peng Cheng Laboratory, Shenzhen, Guangdong, China}
%
%
%
%
\maketitle            
\begin{abstract}
Neurite Orientation Dispersion and Density Imaging (NODDI) is an important imaging technology used to evaluate the microstructure of brain tissue, which is of great significance for the discovery and treatment of various neurological diseases. Current deep learning-based methods perform parameter estimation through diffusion magnetic resonance imaging (dMRI) with a small number of diffusion gradients. These methods speed up parameter estimation and improve accuracy. However, the diffusion directions used by most existing deep learning models during testing needs to be strictly consistent with the diffusion directions during training. This results in poor generalization and robustness of deep learning models in dMRI parameter estimation. In this work, we verify for the first time that the parameter estimation performance of current mainstream methods will significantly decrease when the testing diffusion directions and the training diffusion directions are inconsistent. A robust NODDI parameter estimation method with adaptive sampling under continuous representation (RobNODDI) is proposed. Furthermore, long short-term memory (LSTM) units and fully connected layers are selected to learn continuous representation signals. To this end, we use a total of 100 subjects to conduct experiments based on the Human Connectome Project (HCP) dataset, of which 60 are used for training, 20 are used for validation, and 20 are used for testing. The test results indicate that RobNODDI improves the generalization performance and robustness of the deep learning model, enhancing  the stability and flexibility of deep learning NODDI parameter estimatimation applications.
\keywords{Robustness  \and Diffusion MRI \and Adaptive sampling \and Continuous representation \and NODDI parameter estimation}
\end{abstract}
\section{Introduction}
Diffusion magnetic resonance imaging (dMRI) provides a unique tool for non-invasive assessment of tissue microstructure~\cite{ref_article1}. Neurite Orientation Dispersion and Density Imaging (NODDI) is a popular physiological component-based microstructural model in dMRI~\cite{ref_article2}. NODDI-derived measures can reflect changes in brain microstructural properties across a variety of neurological and psychiatric disorders, as well as brain development, maturation, and aging across the lifespan, including from neonatal to adolescence and adulthood~\cite{ref_article3,ref_article4,ref_article5,ref_article6,ref_article7}.

Advanced dMRI models are highly nonlinear and composed of multiple compartments, so they often require dense sampling in q-space to better estimate parameters. Dense sampling in q-space requires the acquisition of many diffusion-weighted images (DWI) with different diffusion directions and b-value, which is very time-consuming and prone to motion artifacts~\cite{ref_article8}. In the current parameter estimation of advanced dMRI models, optimization-based methods such as nonlinear least squares (NLLS) method, the Markov chain Monte Carlo (MCMC) method, and Bayesian method are easy to produce estimation errors~\cite{ref_article1}. Moreover, deep learning-based methods have been widely used in microstructure estimation research~\cite{ref_article8,ref_article9,ref_article10,ref_article11,ref_article12,ref_article13,ref_article14,ref_article15,ref_article16,ref_article17,ref_article18,ref_article19,ref_article20}. Golkov et al.~\cite{ref_article9} used multilayer perceptron (MLP) for the first time to estimate diffusion kurtosis and microstructural parameters. Next various deep learning networks were used for dMRI model estimation. Chen et al.~\cite{ref_article10,ref_article17,ref_article18,ref_article19} proposed graph neural network (GNN) based structure to estimate the NODDI model parameters. Ye et al. ~\cite{ref_article8,ref_article11,ref_article12,ref_article13} proposed a series of deep learning models to estimate NODDI model parameters. In addition, there are also studies~\cite{ref_article14,ref_article15,ref_article16} linking spherical harmonic (SH) coefficients to dMRI, among which Vishwesh et al.~\cite{ref_article16} directly used SH coefficients to estimate the microstructure. However, this study used images in the full gradient direction of single-shell, did not perform downsampling operations on single-shell, and did not fully exploit the multi-shell signal of dMRI. In summary, deep learning microstructure parameter estimation has achieved good results, speeding up parameter estimation and improving accuracy. However, the current testing process of deep learning models in microstructural parameter estimation needs to be highly consistent with training, that is, fixed DWI gradient direction information needs to be used. This results in poor generalization and clinical applicability of the deep learning model in dMRI microstructural parameter estimation, which does not meet the actual clinical needs.

To overcome these problems, in this article, we propose a robust adaptive sampling NODDI parameter estimation method under continuous representation (RobNODDI). We cut the original DWI into four-dimensional patches, then perform adaptive sampling, and then convert it into a continuous representation signal through SH fitting. Adaptive sampling can ensure that as much data as possible is involved in training, and the model can mine more useful information. Continuous representation helps the model be more flexible when testing diffusion directions and the training diffusion directions are inconsistent. These operations help our model achieve better results in random sampling tests. It is worth noting that our method has no special requirements for model architecture and is theoretically suitable for most current mainstream models~\cite{ref_article21,ref_article22,ref_article23}. Therefore, we selected the advanced MESC-SD as our basic architecture~\cite{ref_article12}.

In this study, our method was verified in the NODDI model. The dataset is the public the Human Connectome Project (HCP) dataset~\cite{ref_article24}. The results show that the proposed method can greatly improve the generalization and robustness of existing deep learning models, and has better clinical applicability.



\section{Method}
RobNODDI combines adaptive sampling and continuous representation. The purpose of adaptive sampling is to improve the network's adaptability to different sampling data and fully mine and use the information in DWI. Continuous representation can make subsequent testing tasks more flexible by converting DWI into SH coefficients and allowing the model to directly learn this continuous representation. These strategies ensure that the diffusion directions of RobNODDI during testing does not need to be consistent with the diffusion directions during training, can effectively estimate parameter results, and significantly improve the generalization~\cite{ref_article25} and robustness of the deep learning model.
\subsection{Overall Architecture}
As shown in Figure 1, our method mainly has two core points: adaptive sampling and continuous representation. We first obtain DWI patches of a total of D diffusion directions for any two shells (bi and bj). During the training stage, we divide multi-shell into two independent shells and perform adaptive sampling respectively. A single shell can be sampled as w×w×w×N, N represents the number of diffusion directions for adaptive sampling, which can be set between 20 and 60. Then the SH coefficients are calculated through linear least squares on the sampled patches, which is a continuously represented signal. Then we concatenate the SH coefficients of the two shells as input to the model for training. The size of the output parameter patch during training is generally set to be smaller than the input patch, and we set it to (w-2)×(w-2)×(w-2)×3. In the testing stage, we can accurately estimate the NODDI parameters by inputting DWI patches of two shells with sizes w×w×w×S1 and w×w×w×S2 respectively. Among them, S1 and S2 can be different from N in training, so RobNODDI will be more flexible and robust in clinical applications.
\begin{figure}
\includegraphics[width=\textwidth]{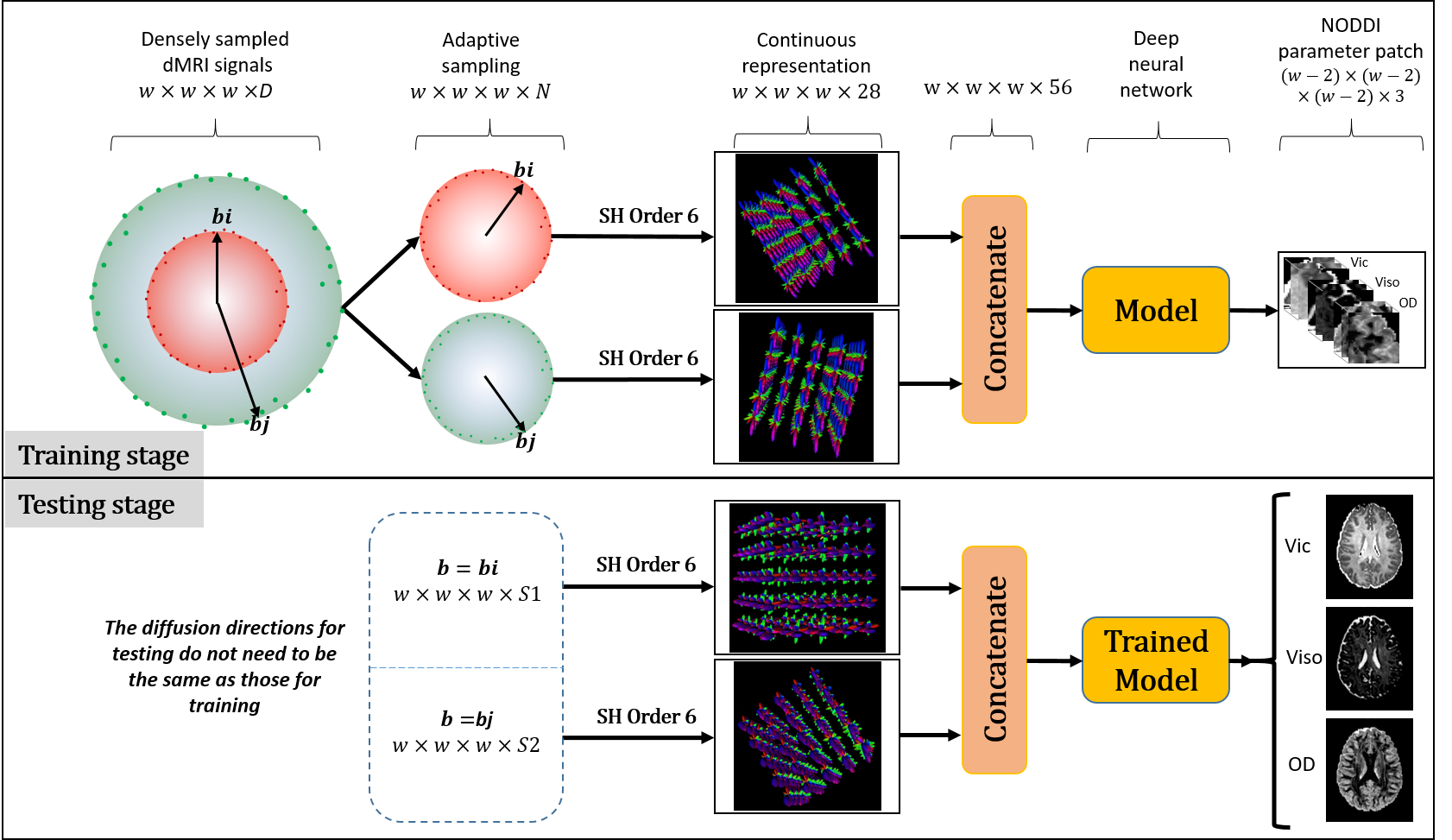}
\caption{Overview of RobNODDI. It contains training stage and testing stage. RobNODDI performs adaptive sampling and SH fitting on the DWI patches, and then concatenates the SH coefficients into the model. Note that both adaptive sampling and SH fitting are included in the training stage, and only SH fitting is included in the testing stage.} \label{fig1}
\end{figure}
\subsection{Continuous representation}
The SH function is a special function defined on the sphere and has orthogonal completeness and continuity. It can be used as a set of orthogonal complete bases for functions on the sphere, and is used to expand and approximate functions on the sphere. Therefore, it is often used in dMRI~\cite{ref_article14,ref_article15,ref_article16,ref_article26,ref_article27}. The expansion formula of dMRI signal on the sphere is shown in formula (1).
\begin{equation}
\mathrm{D}(\theta, \varphi)= {\textstyle \sum_{l=0}^{\infty}} {\textstyle \sum_{m=-l}^{l}}  \hat{C}_{l}^{m} Y_{l}^{m}(\theta, \varphi)
\end{equation}
Here, $\mathrm{D}(\theta, \varphi)$ represents the normalized dMRI signal, $\hat{C}_{l}^{m}$represents the SH coefficient of dMRI,  $Y_{l}^{m}(\theta, \varphi)$represents the SH basis. $l$ and $m$ represent SH order and degree respectively. In dMRI, the direction of the diffusion gradient is described by $\theta, \varphi$. Where $\theta$ and $\varphi$ respectively represent two angles in the spherical coordinate system, namely the polar angle and the azimuth angle. When the basis is formed, the SH coefficients $\hat{C}_{l}^{m}$ can be solved by regularized linear least squares fitting~\cite{ref_article14}. In practical applications, with the original DWI data and corresponding direction coordinate information, we can obtain the corresponding the SH coefficients by solving formula (1).
\subsection{Adaptive sampling}
When sampling DWI signal in dMRI microstructure parameter estimation, uniform sampling is generally performed in q-space~\cite{ref_article11,ref_article26,ref_article27,ref_article28}, which ensures a more balanced use of the information in DWI. But for deep learning models, the sampling of the model during testing must be consistent with the training, which is inflexible and leads to a greatly reduced generalization and robustness of the model.  Therefore, in this study, RobNODDI utilizes a combination of adaptive sampling and continuous representation. Specifically, each DWI patch used for training is randomly sampled, and the sampling pattern varies across different training epochs. Then, we perform Spherical Harmonics (SH) fitting on the adaptively sampled DWI patch, and finally input the processed patch into the network for training. In actual implementation, We first cut the DWI of the original two b values into w×w×w×D, where the fourth dimension D represents all directions of the two arbitrary b values. Then we randomly sample the patches of the two shells separately to obtain patches of size w×w×w×N, where N can be set between 20 and 60. We perform random sampling for each patch to enable the network to achieve adaptive sampling. After training on patches containing N diffusion directions, our model can be tested in multi-shell using S directions per shell that are different from the trained directions. In theory, it has better robustness than existing deep learning methods.
\subsection{Network Construction}
Our proposed method is independent of the network architecture used. Therefore, in this study, we utilize the MESC-SD developed for NODDI used in the study~\cite{ref_article12} as the basic architecture. The network architecture adopts SD-LSTM in the first step, which is constructed by unfolding an iterative process for solving the sparse reconstruction problem, and in the second stage by adding a fully connected layer to calculate the mapping of SH coefficient representations to NODDI parameter.
\section{Experiments}
In this section, we evaluate the proposed method to estimate NODDI parameters, including intracellular volume fraction ($V_{icvf}$), isotropic volume fraction ($V_{iso}$), and directional dispersion ($OD$)~\cite{ref_article2,ref_article12}. The network generalization performance is compared with deep learning methods such as q-DL~\cite{ref_article9}, GCNN~\cite{ref_article10} and MESC-SD~\cite{ref_article12} through SS and RS tests. In particular, we use SS to indicate that the same sampling scheme is used during both testing and training. On the other hand, RS is used to indicate that random sampling is employed during testing, where the diffusion directions in the testing set are different from those used in the training set.
\subsection{Implementation Details}
We used 6-order SH (28 coefficients) in the HCP dataset, with N=30 diffusion directions for each shell, and select two shells of b=1000 s/mm² and b=2000 s/mm². We set w=5, which means cutting the patch into a size of  5×5×5×180, and the corresponding patch size of the NODDI parameter is 3×3×3×3. Additionally, mean square error (MSE) is used as the loss function and Adam is used as the optimizer. The initial learning rate and update method of different experiments are different. The initial learning rate is mainly set to 0.0005 and 0.0001. The training epochs are 30 or 50, and the batchsize of all experiments is set to 128. All models are applied in pytorch2.0 and trained on a server equipped with TITAN Xp GPU. 
\subsection{Dataset and Evaluation Metrics}
\subsubsection{Dataset.}
We selected data from HCP dataset~\cite{ref_article21} to conduct experiments and test the results. The data set contains 3 b values (b=1000, 2000, 3000 s/mm²), 90 diffusion directions for each b value, and an additional 18 non-diffusion weighted volumes. We randomly selected 100 adult subjects and used 60\% of them for training, 20\% for validation, and the remaining 20\% for testing. We obtain the gold standard by fitting all spherical shells with all gradient directions through AMICO~\cite{ref_article29}. 
\subsubsection{Evaluation Metrics.}
We use the root mean square error (MSE), peak signal-to-noise ratio (PSNR), and structural similarity index measure (SSIM) to evaluate the quality of the predicted NODDI-derived indices.
\subsection{Results}
The whole experiment includes:
Testing the poor generalization of current deep learning methods in parameter estimation under different diffusion directions.
Evaluating whether fitting DWI with SH as input can help improve the generalization of existing methods.
Demonstrating that the proposed RobNODDI method achieves the best robustness while ensuring good performance.
\begin{table}[]
\centering
\caption{Average quantitative indicators of NODDI parameters for SS and RS testing using different methods on the q-space data with 30 diffusion directions per shell (1000, 2000 s/mm²). }\label{tab1}
\begin{tabular}{ccccc}
\hline
\multicolumn{1}{c}{Method}         & Sampling in testing & MSE     & PSNR     & SSIM    \\ \hline
\multirow{2}{*}{q-DL {[}9{]}}     & SS                  & 0.00463 & 23.41919 & 0.92024 \\
                                   & RS                  & 0.00851 & 20.74516 & 0.88076 \\ \hline
\multirow{2}{*}{GCNN {[}10{]}}     & SS                  & 0.00405 & 23.97635 & 0.92708 \\
                                   & RS                  & 0.01268 & 19.00210 & 0.86630 \\ \hline
\multirow{2}{*}{MESC-SD {[}12{]}} & SS                  & 0.00219 & 26.64520 & 0.95775 \\
                                   & RS                  & 0.00842 & 20.79057 & 0.89852 \\ \hline
\multirow{2}{*}{MESC-SD with SH}  & SS                  & 0.00226 & 26.49617 & 0.95626 \\
                                   & RS                  & 0.00263 & 25.82884 & 0.94980 \\ \hline
\multirow{2}{*}{RobNODDI}          & SS                  &\textbf{0.00211}  & \textbf{26.80304} & \textbf{0.95901} \\
                                   & RS                  & \textbf{0.00211} & 26.79317 & 0.95884 \\ \hline
                                   
\end{tabular}
\end{table}
\subsubsection{SS and RS tests of existing deep learning methods.}
Table 1 shows the quantitative metrics of three deep learning methods under SS and RS testing. It can be observed that the existing deep learning methods suffer a significant performance drop when the diffusion direction changes, making it difficult to estimate the parameters accurately. The visual comparison in Figure 2 also demonstrates that the existing deep learning methods exhibit a clear performance degradation when the testing diffusion direction is altered. In other words, most existing deep learning methods have poor generalization to changes in dMRI diffusion direction.
\begin{figure}
\includegraphics[width=\textwidth]{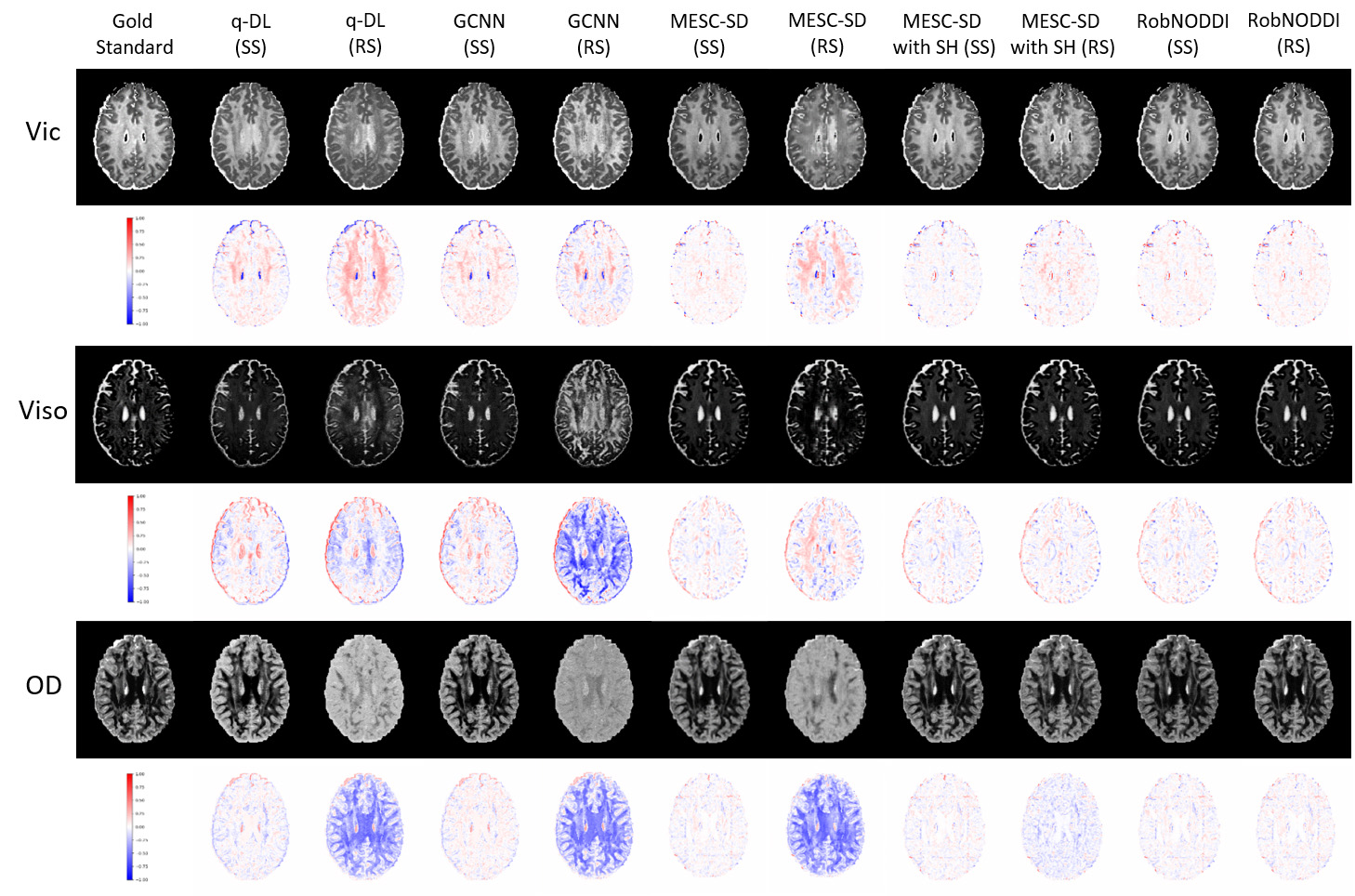}
\caption{Qualitative comparison of NODDI parameter for SS and RS testing using different methods with 30 diffusion directions per shell (1000, 2000 s/mm²)} \label{fig2}
\end{figure}
\subsubsection{SS and RS testing using SH coefficients as input.}
The input for training here is the SH coefficient of 30 diffusion directions with fixed uniform sampling per shell. The test is also divided into two types: SS and RS. The results are shown in Figure 2 and Table 1 corresponding to MESC-SD with SH (SS) and MESC-SD with SH (RS). It can be seen that using SH as input to estimate microstructure parameters is effective, and the results are very close to using DWI as input. When the test direction is RS, the performance also decreases; however, compared to the original MESC-SD, the RS test shows much better results. This indicates that using continuous representation of DWI as input improves the model's generalization performance in dMRI microstructure parameter estimation, yet it still lags behind the SS test. Specifically, for Vic and OD, the visual results and error maps in Figure 2 still show some discrepancy compared to the SS testing results.
\subsubsection{SS and RS test of the proposed RobNODDI method.}
This part introduces our proposed RobNODDI method. As shown in Figure 1, our input is the DWI patch, and we incorporate adaptive sampling and SH coefficient fitting into the network training. During testing, we also conducted SS and RS evaluations for RobNODDI, where the SS testing uses the same sampling scheme as the comparative methods. The corresponding results are shown in Figure 2 and Table 1. It can be observed that when the testing direction changes, the results estimated by our method remain highly consistent, with almost no visible differences, and the quantitative metrics are very close as well. Compared to the original MESC-SD method, our approach achieves significantly improved generalization. Moreover, compared to MESC-SD with SH, the proposed method further enhances both performance and generalization. Therefore, our method not only achieves more accurate parameter estimation, but also demonstrates strong generalization and robustness.
\subsection{Ablation Study}
\subsubsection{RS test of RobNODDI method with different number of samples.}
\begin{table}[]
\centering
\caption{RobNODDI performs RS test of NODDI parameters on q-space data with different number of diffusion directions (1000, 2000 s/mm²)}\label{tab2}
\begin{tabular}{cccccc}
\hline
Method                    & Sampling in testing & Number of directions & MSE     & PSNR     & SSIM    \\
\hline
\multirow{7}{*}{RobNODDI} & \multirow{7}{*}{RS} & 20/20                   & 0.00231 & 26.40777 & 0.95512 \\ \cline{3-6}
                          &                     & 25/25                   & 0.00217 & 26.67221 & 0.95759 \\ \cline{3-6}
                          &                     & 30/30                   & 0.00211 & 26.80304 & 0.95901 \\ \cline{3-6}
                          &                     & 35/35                   & 0.00208 & 26.85303 & 0.95952 \\ \cline{3-6}
                          &                     & 40/40                   & 0.00208 & 26.87385 & 0.95989 \\ \cline{3-6}
                          &                     & 16/29                   & 0.00226 & 26.50507 & 0.95590 \\ \cline{3-6}
                          &                     & 21/28                   & 0.00219 & 26.64630 & 0.95730 \\ \cline{3-6}
                          &                     & 26/23                   & 0.00219 & 26.64381 & 0.95732 \\ \hline
\end{tabular}
\end{table}

To further validate the robustness of the proposed method, we conducted ablation experiments on RobNODDI by randomly sampling DWI with varying diffusion directions and numbers. This included tests with consistent numbers of diffusion directions for two b values: 40, 50, 70, and 80 directions (20, 25, 35 and 40 for b = 1000 s/mm² and b = 2000 s/mm²). And tests with different numbers of diffusion directions for two b values: 45, 49, and 49 directions (16, 21 and 26 for b = 1000 s/mm²; 29, 28 and 23 for b = 2000 s/mm²). The ablation results are presented in Table 2, highlighting the robustness of RobNODDI.

It is evident that performance slightly improves with an increasing number of diffusion directions used in testing. Moreover, our method consistently maintains stable performance and high flexibility across different numbers of DWI signals used for testing.
\section{Conclusion and Future Work}
In this work, we propose RobNODDI, which achieves robust NODDI parameter evaluation by combining adaptive sampling and continuous representation. By comparing with the existing deep learning microstructure estimation methods, it is shown that converting dMRI images into SH coefficients helps to improve the generalization of the original method for diffusion directions. The proposed RobNODDI can further improve the robustness and stability of the deep learning model in dMRI microstructure parameter estimation, making the deep learning model more flexible and universal in dMRI clinical applications. Moreover, the model architecture in the method does not rely on a specific network structure.
In our future work, we will explore more advanced methods, such as~\cite{ref_article15,ref_article30}, and further optimize the proposed approach.
\subsubsection{\ackname} This research was partly supported by the National Natural Science Foundation of China (62222118, U22A2040, 62171261, 81671848, 81371635), Guangdong Provincial Key Laboratory of Artificial Intelligence in Medical Image Analysis and Application (2022B1212010011), Shenzhen Science and Technology Program (RCYX20210706092104034, JCYJ20220531100213029), Key Laboratory for Magnetic Resonance and Multimodality Imaging of Guangdong Province (2023B1212060052), Youth lnnovation Promotion Association CAS, National Key R\&D Program of China (2023YFA1011400), Fundamental Research Funds for the Central Universities (China), and Innovation Ability Improvement Project of Science and Technology Small and Medium-sized Enterprises of Shandong Province (2021TSGC1028).
%
%
%
%

\end{document}